\documentclass{article} % For LaTeX2e
\usepackage{nips15submit_09,times}
\usepackage{graphicx}
\usepackage{hyperref}
\usepackage{wrapfig}
\usepackage{url}
\usepackage{amsmath}
\usepackage{amsmath,amssymb}

\DeclareMathOperator*{\argmax}{arg\,max}

\title{Q-Networks for Binary Vector Actions\thanks{This paper was accepted at Deep Reinforcement Learning Workshop, NIPS 2015}}

\author{
Naoto Yoshida\\
Tohoku University \\
Aramaki Aza Aoba 6-6-01\\
Sendai 980-8579, Miyagi, Japan\\
\texttt{naotoyoshida@pfsl.mech.tohoku.ac.jp} \\
}

\nipsfinalcopy % Uncomment for camera-ready version

\begin{document}

\maketitle

\begin{abstract}
In this paper reinforcement learning with binary vector actions was investigated. We suggest an effective architecture of the neural networks for approximating an action-value function with binary vector actions. The proposed architecture approximates the action-value function by a linear function with respect to the action vector, but is still non-linear with respect to the state input. We show that this approximation method enables the efficient calculation of greedy action selection and softmax action selection. Using this architecture, we suggest an online algorithm based on Q-learning. The empirical results in the grid world and the blocker task suggest that our approximation architecture would be effective for the RL problems with large discrete action sets. 
\end{abstract}

\section{Introduction}
One of the big challenges in reinforcement learning (RL) is learning in high dimensional state-action spaces. Recent advances in deep learning technologies have enabled us to treat RL problems with the high-dimensional state space, and it achieved an impressive result in general game playing tasks ({\it e.g.} ATARI game plays) \cite{mnih2015human}. 

Even though several approaches are suggested for RL with continuous actions \cite{lillicrap2015continuous}\cite{theodorou2010reinforcement}, RL with a large action space is still problematic, especially when we treat binary vectors as representations of the actions. The difficulty is that the number of actions exponentially grows as the length of the binary vector grows. Recently, several approaches have been used to tackle this problem. Sallans \& Hinton suggested an energy-based approach in which restricted Boltzmann machines \cite{smolensky1986information} were adopted in the algorithm and their free energy was used as the function approximator \cite{sallans2000using}. Heess {\it et al.} followed their energy-based approach and investigated natural actor-critic algorithms with energy-based policies by RBMs \cite{heess2012actor}. Although energy-based approaches are known to be effective in large discrete domains, exact action sampling is intractable due to the nonlinearity of the approximation architecture. Hence, an energy-based approach samples actions by Gibbs sampling. 

However, the Gibbs sampling-based action selection is computationally expensive and requires careful tuning of the parameters. Also, because of the intractability of the exact sampling of greedy actions, no Q-learning-based online off-policy RL algorithm has so far been proposed for the large discrete action domain. From this background, we treat this issue and suggest novel architecture for the off-policy RL with the a large discrete action set.

\section{Preliminaries}
\subsection{Markov Decision Process and Reinforcement Learning}
The value-based reinforcement learning algorithms utilize the Markov decision process (MDP) assumption. The MDP is defined by a tuple $\langle {\cal S}, {\cal A}, P, R\rangle$. ${\cal S}$ is the state set, ${\cal A}$ is the action set, $P$ is the transition probability $P(s'|s,a)$, where  $s' \in {\cal S}$ is the next state given a state-action pair $(s,a)$. Finally $R$ is the average reward function $R(s,a) = \mathbb{E}[r| s,a]$ and $r$ is the reward sample. 

In the value-based RL, the action-value function $Q^\pi(s,a)$ is defined by
\begin{eqnarray}
Q^\pi(s,a) = {\rm E}_\pi \Bigl[ \sum_{t=0}^\infty \gamma^t r_t \Bigl| s_0 = s, a_0 = a \Bigr],
\end{eqnarray}
here, $0 \leq \gamma < 1$ is the discount factor. In value-based RL, we look for the optimal policy $\pi^*$ that maximizes the action-values for every state-action pair. Q-learning is an algorithm for finding the optimal policy in MDP \cite{watkins1989learning}, and the advantage of Q-learning is its off-policy property: the agent can directly approximate the action-value of an optimal policy $\pi^*$ while following the other policy $\pi$.  

Although Q-learning is guaranteed to approximate optimal action-values when we use the tabular functions in a discrete state-action environment \cite{watkins1992q}, tabular function-based approaches become quickly inefficient for RL with large state-action spaces. Then, function approximations become necessary in such domains.

\subsection{Q-learning with Function Approximation}
In the Q-learning algorithm with function approximations, we approximate the optimal value function by the function $Q_\theta(s,a)$ where $\theta$ is the parameter of the function.

The gradient-based update of the function $Q_\theta(s,a)$ calculates the gradient of the error function 
\begin{eqnarray}
L = \frac{1}{2}(T - Q_\theta(s, a))^2,
\end{eqnarray}
where $T$ is the target signal. Then, the gradient of the error is obtained by
\begin{eqnarray}
\frac{\partial L}{\partial \theta} = -(T - Q_\theta(s, a)) \frac{\partial  Q_\theta(s, a)}{\partial \theta}.
\end{eqnarray}
The target signal in the Q-learning is $T = r + \gamma \max_{\hat a} Q_\theta(s', \hat a)$ given a transition sample $\{s,a, r, s'\}$. Then the direction of the parameter update ${\it \Delta}\theta$ is given by
\begin{eqnarray}
{\it \Delta}\theta &=& - \frac{\partial L}{\partial \theta}\\
&=& \Bigl(r + \gamma \max_{\hat a} Q_\theta(s', \hat a) - Q_\theta(s, a)\Bigr) \frac{\partial  Q_\theta(s, a)}{\partial \theta}.
\end{eqnarray}
The first term of the product in the second equality is called the TD error. Using this gradient, the stochastic gradient descent or more sophisticated gradient-based algorithms are used for approximating the optimal action-value function \cite{lin1992self}\cite{mnih2013playing}.
 
The Q-learning-based gradient requires the max operation of $Q_\theta(s, a)$ given a state. In the previous research with small discrete action sets, this max operation were tractable. However, if the actions are composed of binary vectors or factored representation \cite{sallans2000using}\cite{sallans2004reinforcement}, the number of total actions exponentially grows and quickly become intractable.

\section{Proposed Method}
\begin{figure}[t] 
  \begin{center}
    \includegraphics[width = 5.0cm]{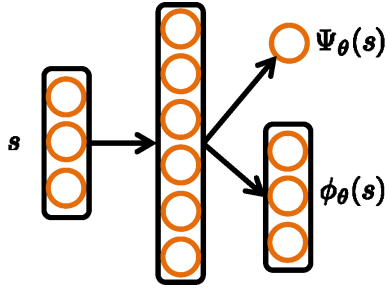}
  \end{center}
    \caption{An example of the network architecture of the proposed method.}
    \end{figure}

In this study, we assume that the function approximation is done by the multi-layer perceptrons (MLPs)  parameterized by $\theta$. To efficiently calculate the max operations in Q-learning with a large discrete action space, we propose the network architecture of MLPs shown in Figure 1. In this architecture, the outputs of the network are composed of a continuous scalar variable $\Psi_\theta(s)$ and continuous vector variable $\phi_\theta(s)$. In this study, we approximate the action-value function by the linear function with respect to the action vector:
\begin{eqnarray}
Q_\theta(s,a) &=& \Psi_\theta(s) + \sum_{i=1}^K a_i \phi^i_\theta(s)\\
&=& \Psi_\theta(s) + a^\top \phi_\theta(s)
\label{eq_action_linear}
\end{eqnarray}
Here, $a$ is the action represented by the binary vector, and $a_i$ is the $i$-th component of the action. 

The gradient of the function $Q_\theta(s, a)$ is given by
\begin{eqnarray}
\frac{\partial  Q_\theta(s, a)}{\partial \theta} = \frac{\partial   \Psi_\theta(s)}{\partial \theta} + \sum_{i=1}^K a_i \frac{\partial \phi^i_\theta(s)}{\partial \theta},
\label{eq_gradient}
\end{eqnarray}
and this is efficiently obtained by the back propagation algorithm.  

\subsection{Sampling of the Actions}
The proposed approximation architecture provides an efficient calculation of the greedy action. For actions with the one-hot representation, the greedy policy is obvious. This is
\begin{eqnarray}
\pi_{\rm greedy}(s) &=& \argmax_{a\in \{1,...K\}} Q_\theta(s,a)\\
&=& \argmax_{i\in\{1,...K\}} \phi_\theta^i(s),
\end{eqnarray}
where $\phi_\theta^i(s)$ is the $i$-th element of the outputs $\phi_\theta(s)$. 

For the $K$-bits binary vector actions, sampling of the greedy actions with respect to the function \ref{eq_action_linear} is still tractable. The $i$-th element of the greedy action vector is given by
\begin{eqnarray}
\pi^i_{\rm greedy}(s) = 
\begin{cases}
    0 &  \text{$\phi^i_\theta(s) < 0$}\\
    1 &  \text{otherwise}.
\end{cases}
\end{eqnarray}
Because we can efficiently sample the greedy action, the $\epsilon$-greedy action selection is tractable in our case. In the experiment section, we tested some variants of the $\epsilon$-greedy action selection.

The exact sampling from the softmax action selection for binary vector actions is also tractable. Substituting the equation \ref{eq_action_linear} into the conventional softmax policy with the inverse temperature $\beta>0$ gives the equality
\begin{eqnarray}
\pi(a|s) &=& \frac{e^{\beta Q_\phi(s,a)}}{\sum_{a'\in{\cal A}} e^{\beta Q_\phi(s,a')}}\\
&=& \frac{e^{\beta \sum_{i=1}^K a_i \phi^i_\theta(s)}}{\sum_{a'\in{\cal A}} e^{\beta \sum_{i=1}^K a'_i \phi^i_\theta(s)}}\\
&=& \prod_{i=1}^K  \frac{e^{\beta a_i \phi^i_\theta(s)}}{\sum_{a'_i\in\{0, 1\}} e^{\beta  a'_i \phi^i_\theta(s)}}\\
&=& \prod_{i=1}^K \pi_i (a_i|s),
\label{eq_binary}
\end{eqnarray}
where $\pi_i(a_i|s)$ is the bernoulli distribution for the $i$-th element of the action. The firing probability of the $i$-th bit of the action is given by the logistic function 
\begin{eqnarray}
\pi_i(a_i = 1|s) &=& \frac{1}{1 + e^{-\beta \phi^i_\theta(s)}}.
\end{eqnarray}

When the environment is represented by the factored MDP \cite{sallans2000using}\cite{sallans2004reinforcement}, the action may be represented by the binary vector, which is composed of a concatenation of one-hot representation vectors (for example, the agent may have to decide one of 2 options and one of 3 options simultaneously. In this case, if the agent takes the first option and third option, an action is represented as a 5-bit vector $(1, 0|\ 0, 0, 1)^\top$). The greedy action for the factored environment is given by
\begin{eqnarray}
\pi^j_{\rm greedy}(s) &=&  \argmax_{i\in\{1,...K_j\}} \phi_\theta^{ij}(s),
\end{eqnarray}
where $j$ is the index of the factored action sets, and $K_j$ is the size of the $j$-th action set. Following a similar transformation of the equation \ref{eq_binary}, the softmax policy for the factored action is given as
\begin{eqnarray}
\pi(a|s) = \prod_{j=1} \pi_j (a^j|s),
\end{eqnarray}
and $ \pi_j (a^j|s)$ is the softmax function with respect to the $j$-th factored action set
\begin{eqnarray}
\pi_j(a_i^j = 1|s) &=& \frac{e^{\beta \phi^{ij}_\theta(s)}}{\sum_{i=1}^{K_j} e^{\beta  \phi^{ij}_\theta(s)}}.
\end{eqnarray}

\section{Experiment}
In the experiment, we tested our proposed architecture in several domains. In all of the experiments, we used the three-layer MLPs described in Figure 1. We also set the activation function of the hidden units using the rectifier linear units (ReLU). All weights connected with output units are sampled from the uniform distribution over $[-0.01, 0.01]$, and all weights between input units and hidden units are sampled from the uniform distribution over $[-\sqrt{6}/\sqrt{N_{\rm hidden} + N_{\rm input}}, \sqrt{6}/\sqrt{N_{\rm hidden} + N_{\rm input}}]$, where $N_{\rm hidden}$ and $N_{\rm input}$ are the number of units in the layers. The update of the parameter was done by the stochastic gradient descent with a constant step size $\alpha=0.01$. The discount rate of the objective function in RL is also same in the all of the experiments, so we used $\gamma = 0.95$. 

\begin{figure}[t]
  \begin{center}
    \begin{tabular}{c}

      % 3
      \begin{minipage}{0.5\hsize}
        \begin{center}
          \includegraphics[clip, width=5cm]{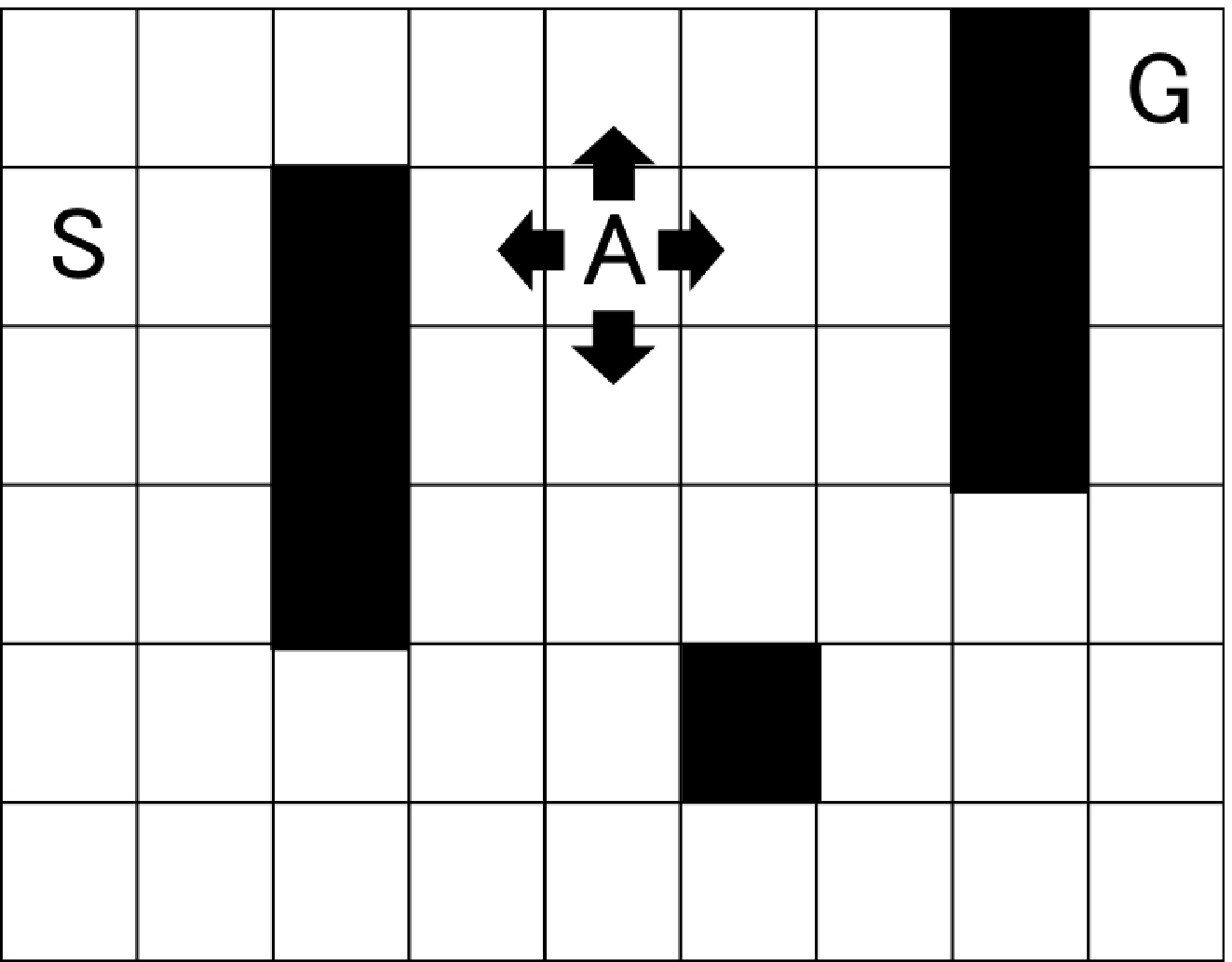}
        \end{center}
      \end{minipage}

      \begin{minipage}{0.5\hsize}
        \begin{center}
          \includegraphics[clip, width=6cm]{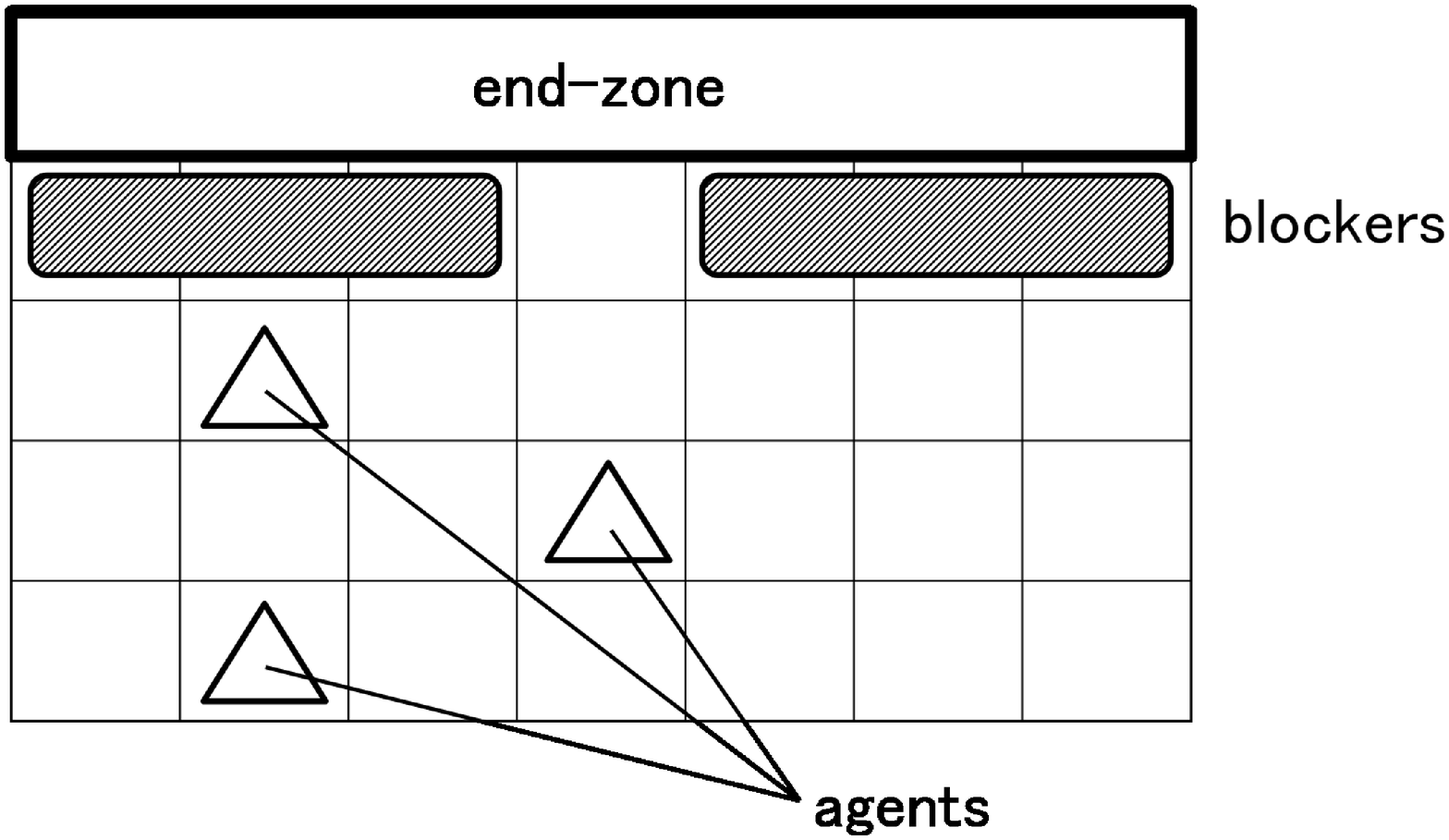}
        \end{center}
      \end{minipage}
      
    \end{tabular}
    \caption{Environments: Sutton's grid world (left) and Blocker (right)}
    \label{fig_results}
  \end{center}
\end{figure}

\subsection{Grid World with One-hot Representation}
First, we tested our algorithm in the conventional grid world with a one-hot representation. This task is the shortest-path problem in the grid world, as suggested by Sutton \& Barto \cite{sutton1998reinforcement}. The state space is composed of 47 discrete states and they are given by the one-hot representation. The agent has 4 discrete actions that correspond to the 4 direction moves (North, South, East, West). The action in this experiment is represented by a one-hot representation (for example, the ``North" action corresponds to the vector $(1,0,0,0)^\top$). The agent receives a zero reward when the agent reaches the goal, but otherwise receives a $-1$ reward.  The agent was trained in an episodic manner, a single episode was terminated when the agent reached the goal or passed 800 time steps in the episode. The agents were implemented by MLPs with 50 hidden units. The $\epsilon$-greedy policy was used as the behavior policy. In this task, we used $\epsilon = 0.1$.

The left panel of Figure \ref{fig_results} is the result of the experiment. The horizontal axis represents the number of episodes, the vertical axis is the step size in the episode. The black line is the mean performance of 10 runs and the bars are standard deviations. The broken line is the optimal step size. As expected, the agent successfully obtained the optimal policy.

\subsection{Grid World with 4-bit Binary Vector Actions}
\begin{wraptable}{l}{4.5cm}
%\begin{table}[t]
\renewcommand{\arraystretch}{1.3}
\caption{Binary Vector Actions}
\label{table_vector_action}
\centering
  \begin{tabular}{|l|c|r||r|} \hline
    Action & Binary Vector \\ \hline
    North & 1,1,0,0 \\
    South & 0,0,1,1 \\
    East & 1,0,1,0 \\
    West & 0,1,0,1 \\
    Stay & $\rm otherwise$ \\\hline
  \end{tabular}
%\end{table}
\end{wraptable}
In this environment, the task is also the shortest-path problem in the same grid world. The state is given as a one-hot representation, as well. The agent receives a zero reward when it reaches the goal, but otherwise receives a $-1$ reward. The training is  episodic and the termination rule of a single episode is the same as in the previous experiment. In this experiment, actions are represented by 4-bit binary vectors as shown in Table \ref{table_vector_action}. Only 4 of $2^4 = 16$ patterns move the agent to the corresponding direction, and the agent stays at the same state if the other action patterns are selected. The agents were again implemented by MLPs with 50 hidden units. The $\epsilon$-greedy policy was used as the behavior policy. In this task, we used $\epsilon = 0.2$.

Right panel of Figure \ref{fig_results} The horizontal axis represents the number of episodes, the vertical axis is the step size in the episode. The black line is the mean performance of 10 runs and the bars are standard deviations. The broken line is the optimal step size. Again, the agent successfully obtained the optimal policy through the experiment even in the binary vector action domain. This result shows that the proposed method successfully improved the behavior of the agent without any Monte-Calro based samplings of the actions, even when the representation of actions is not a one-hot representation.

\begin{figure}[t]
  \begin{center}
    \begin{tabular}{c}
      \begin{minipage}{0.5\hsize}
        \begin{center}
          \includegraphics[clip, width=6cm]{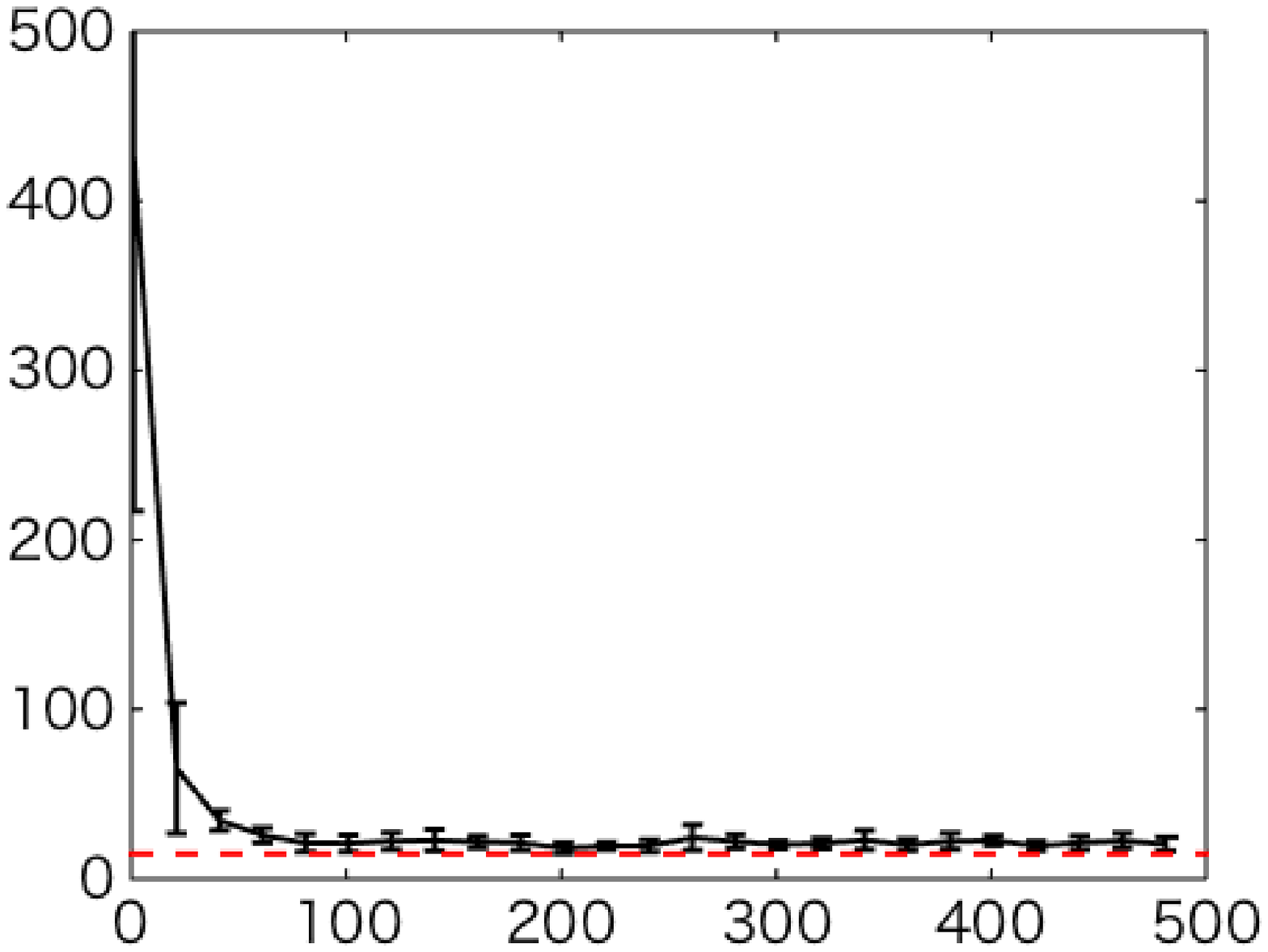}
        \end{center}
      \end{minipage}
      \begin{minipage}{0.5\hsize}
        \begin{center}
          \includegraphics[clip, width=6cm]{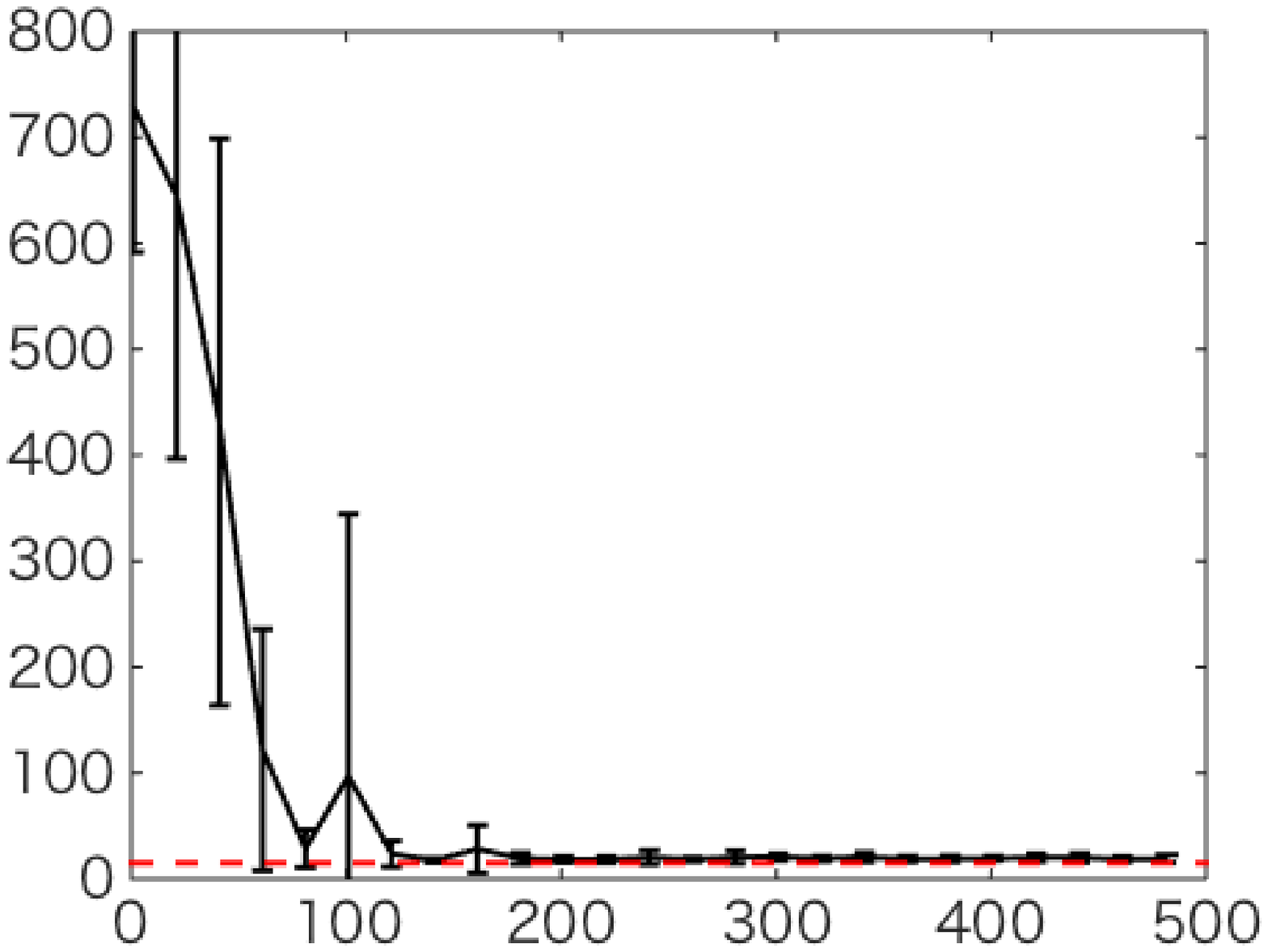}
        \end{center}
      \end{minipage}
    \end{tabular}
    \caption{Results in Sutton's grid world. The broken line represents the minimum time step. Left: grid world; Right: grid world with 4-bit binary vector actions.}
    \label{fig_results}
  \end{center}
\end{figure}

\subsection{Grid World with Population Coding}
Again, the task is the shortest path problem in the same grid world. The state representation, the reward function and termination rules of a single episode are the same as in the  previous experiments. In this experiment, the action is represented by a 40-bit binary vector. And the moves of the agent are driven according to the type of population coding. Concretely, when the environment receives a 40-bit vector, one of the four-direction moves (1: North; 2: South; 3: East; 4: West) or the stay behavior (5: Stay) occurs according to the probability
\begin{eqnarray}
P_j = \frac{E_j}{\sum_{k=1}^5 E_k}&  j =1,2,3,4,5,
\end{eqnarray}
where $E_j$ are give by the action $a_i \in \{0, 1\},\ i = 1,\dots,40$ following the equations
\begin{eqnarray}
E_1 = {\sum_{i=1}^{10} a_i},\ E_2 = {\sum_{i=11}^{20} a_i},\ E_3 = {\sum_{i=21}^{30} a_i},\ E_4 = \sum_{i=31}^{40} a_i
\end{eqnarray}
and
\begin{eqnarray}
E_5 = \max \Bigl(10 - \sum_{k=1}^4 E_k,\ 0\Bigr).
\end{eqnarray}
In this experiment, because the discrete action space exponentially grows according to the length of the binary action vector, the size of the corresponding action space is huge $|{\cal A}| = 2^{40} > 10^{12}$. Therefore, efficient sampling of the action is also required in this domain. 

In this experiment, we used MLPs with 50 hidden units. We tested three types of behavior policies. The first policy is the conventional $\epsilon$-greedy policy. We used $\epsilon = 0.3$ in the task. the second policy is the bit-wise $\epsilon$-greedy policy. In this policy, each bit of the action element undertakes $\epsilon_{\rm bit}$-greedy exploration. More concretely, the $i$-th element of the action vector takes the random action ($a_i=1$ with probability 0.5) with probability $\epsilon_{\rm bit}$. Because we can sample the greedy actions with ease, we can explicitly take this behavior policy. We used $\epsilon_{\rm bit}=0.05$ in this experiment. The third policy is the sofmax policy that was explained in section 3.1. We used $\beta = 20$ in this experiment.

\begin{figure}[t] 
  \begin{center}
    \includegraphics[width = 13.0cm]{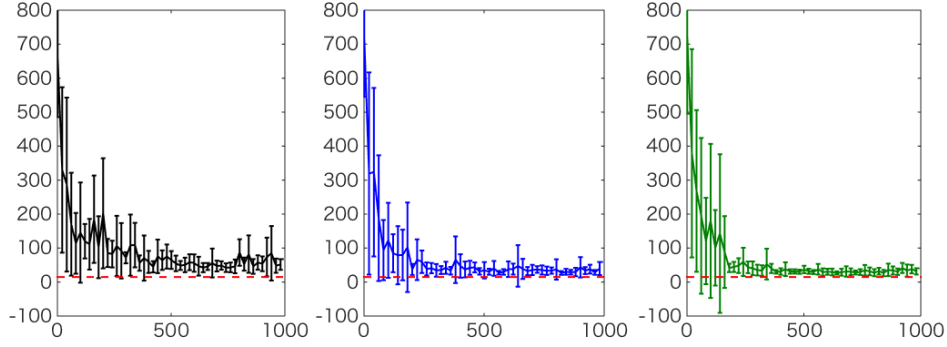}
  \end{center}
    \caption{The result of the grid world with population coding task. The left most panel shows the result of the $\epsilon$-greedy action selection with $\epsilon=0.3$, the center panel is that of the bit-wise $\epsilon$-greedy with $\epsilon_{\rm bit} = 0.05$. The right most panel is the result of the softmax action selection with the inverse temperature $\beta=20$.}
    \label{result_pop}
    \end{figure}

Figure \ref{result_pop} shows the result of the experiment. The horizontal axis represents the number of episodes, the vertical axis is the step size in the episode. The solid lines are the mean performance of 10 runs and the bars are standard deviations. The broken lines are the optimal step size. The results show that all three behavior policies successfully improved the performance of the agent in the high-dimensional action space. From these results, the bit-wise $\epsilon$-greedy policy (center) and the softmax policy (right) shows better performance than that of the conventional $\epsilon$-greedy policy (left). This would be because of the large exploration rate in the $\epsilon$-greedy policy ($\epsilon=0.3$), but running with a smaller exploration rate ($\epsilon \leq 0.2$) sometimes resulted in divergence of the parameters during the learning. 

\subsection{ Blocker }
The blocker is the multi-agent task suggested by Sallans and Hinton \cite{sallans2000using}\cite{sallans2004reinforcement}. This environment consists of a $4\times7$ grid, three agents, and two pre-programmed blockers. Agents and blockers never overlap each other in the grid. To obtain a positive reward, agents need to cooperate in this environment. The ``team" of agents obtain a $+1$ reward when any one of the three agents enters the end-zone, otherwise the team receives a $-1$ reward. The state vector is given as a 141 binary vector, composed of the positions (grid cells) of all the agents (28 bits $\times$ 3 agents), the east most positions of each blocker (28 bits $\times$ 2 blockers) and a bias bit that is always one (1 bit). Each agent can move to any of the four directions. Hence the size of the action space is $4^3 = 64$. In this environment, the representation of the action is given as a 12-bit binary vector in which the three one-hot representation is concatenated (for example, (North, North, North) actions corresponding to the vector $(1,0,0,0| 1,0,0,0| 1,0,0,0)^\top$).  In each episode, the agents start at a random position in the bottom row of the grid. When one of the agents enters the end-zone or 40 time steps have passed, the episode terminates and the next episode starts after the initialization of the environment.

In this task, we used MLPs with 100 hidden units. The $\epsilon$-greedy policy was used as the behavior policy. In this task, we used $\epsilon = 0.3$. Also, we tested the agent-wise $\epsilon$-greedy policy as the behavior policy. This policy is a modified version of the $\epsilon$-greedy policy for actions with factored representation, and each agent follows the $\epsilon$-greedy policy independently. In the case of the agent-wise-$\epsilon$-greedy policy, we used $\epsilon=0.1$ for each agent.

\begin{figure}[t] 
  \begin{center}
    \includegraphics[width = 11.0cm]{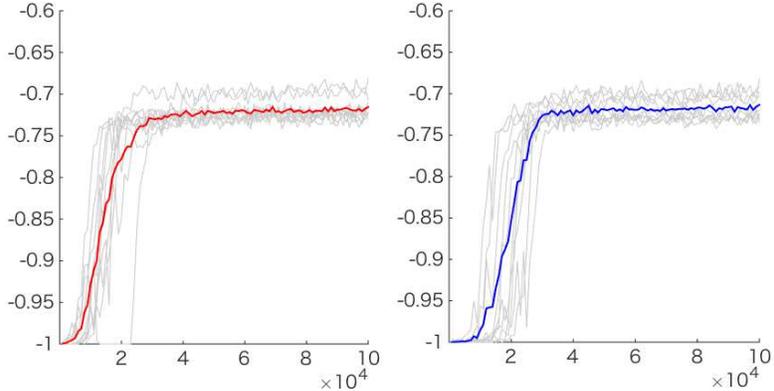}
  \end{center}
    \caption{The result of the blocker task. The left panel is the result of the conventional $\epsilon$-greedy action selection with $\epsilon = 0.3$. The right panel is that of the agent-wise $\epsilon$-greedy action selection with $\epsilon_{\rm agent} = 0.1$. The gray lines are the results of 10 individual runs, and thick lines (red and blue) are the average performance.}
    \label{blocker_result}
    \end{figure}

Figure \ref{blocker_result} shows the results of the experiment. The horizontal axis represents the time steps, and the vertical axis represents the average reward during the last 1000 steps. The left panel is the result of the conventional $\epsilon$-greedy action selection, the right panel is that of the agent-wise $\epsilon$-greedy action selection. Both results are competitive, but in this experiment, agent-wise $\epsilon$-greedy agents tend to escape from the local optima. 

%\begin{figure}[t]
%  \begin{center}
%    \begin{tabular}{c}
%
%      % 3
%      \begin{minipage}{0.5\hsize}
%        \begin{center}
%          \includegraphics[clip, width=6cm]{figure/result_blocker.eps}
%          %\hspace{1.6cm} Blocker
%        \end{center}
%      \end{minipage}
%
%    \end{tabular}
%    \caption{Results in blocker}
%    \label{fig_results}
%  \end{center}
%\end{figure}

\section{Discussion}
In the environment with one-hot representation actions, the linear function approximation of the action-value corresponds to the bilinear function with respect to the action vector and the state vector
\begin{eqnarray}
Q_\theta(s,a) = a^\top \theta s.
\end{eqnarray}
In this case, the parameter $\theta$ is give as a matrix. If the state is given by a one-hot representation, this approximation is identical with the table representation. As suggested in our method, the linear architecture with respect to the action enables efficient sampling of the greedy action.  More recently, Mnih {\it et al.} proposed a DQN architecture \cite{mnih2013playing}. In this case, we evaluate the action-values corresponding to all the discrete actions by a single forward propagation. And then the training of the approximator is done only on the output, which corresponds to the selected action. This architecture can be interpreted as a linear function approximation with respect to the actions 
\begin{eqnarray}
Q_\theta(s,a) =  a^\top \phi_\theta(s).
\end{eqnarray}
If we construct $\phi_\theta(s)$ by some nonlinear function with high representational power such as deep neural networks, this approximation is sufficient for approximating the Q-values when actions are given by one-hot representation vectors.

The goal of our architecture (equation \ref{eq_action_linear}) is to adapt these ideas to the RL with binary vector actions. Although our function approximator is strongly restricted by the linear architecture with respect to the action, our function approximator is sufficient to represent an arbitrary deterministic policy $\pi(s)$ by $\argmax_a Q_\theta(s,a)$ even when we treat the binary vector actions, as long as we represent $\phi_\theta(s)$ by a universal function approximator. 

\section{Conclusion}
In this paper, we suggest a novel architecture of multilayer perceptrons for RL with a large discrete action set. In our architecture, the action-value function is approximated by a linear function with respect to the vector actions. This approximation method enables us to efficiently sample from the greedy policy and the softmax policy. The Q-learning-based off-policy algorithm is therefore tractable in our architecture without any Monte-Carlo approximations. We empirically tested our method in several discrete action domains, and the results supported its effectiveness. Based on these promising results, we expect to extend our approach using deep architectures in a future work. 

\bibliographystyle{unsrt}
\bibliography{reference}{}

\begin{thebibliography}{10}

\bibitem{mnih2015human}
Volodymyr Mnih, Koray Kavukcuoglu, David Silver, Andrei~A Rusu, Joel Veness,
  Marc~G Bellemare, Alex Graves, Martin Riedmiller, Andreas~K Fidjeland, Georg
  Ostrovski, et~al.
\newblock Human-level control through deep reinforcement learning.
\newblock {\em Nature}, 518(7540):529--533, 2015.

\bibitem{lillicrap2015continuous}
Timothy~P Lillicrap, Jonathan~J Hunt, Alexander Pritzel, Nicolas Heess, Tom
  Erez, Yuval Tassa, David Silver, and Daan Wierstra.
\newblock Continuous control with deep reinforcement learning.
\newblock {\em arXiv preprint arXiv:1509.02971}, 2015.

\bibitem{theodorou2010reinforcement}
Evangelos Theodorou, Jonas Buchli, and Stefan Schaal.
\newblock Reinforcement learning of motor skills in high dimensions: A path
  integral approach.
\newblock In {\em Robotics and Automation (ICRA), 2010 IEEE International
  Conference on}, pages 2397--2403. IEEE, 2010.

\bibitem{smolensky1986information}
P~Smolensky.
\newblock Information processing in dynamical systems: foundations of harmony
  theory.
\newblock In {\em Parallel distributed processing: explorations in the
  microstructure of cognition, vol. 1}, pages 194--281. MIT Press, 1986.

\bibitem{sallans2000using}
Brian Sallans and Geoffrey~E Hinton.
\newblock Using free energies to represent q-values in a multiagent
  reinforcement learning task.
\newblock In {\em NIPS}, pages 1075--1081, 2000.

\bibitem{heess2012actor}
Nicolas Heess, David Silver, and Yee~Whye Teh.
\newblock Actor-critic reinforcement learning with energy-based policies.
\newblock In {\em EWRL}, pages 43--58. Citeseer, 2012.

\bibitem{watkins1989learning}
Christopher John Cornish~Hellaby Watkins.
\newblock {\em Learning from delayed rewards.}
\newblock PhD thesis, University of Cambridge, 1989.

\bibitem{watkins1992q}
Christopher~JCH Watkins and Peter Dayan.
\newblock Q-learning.
\newblock {\em Machine learning}, 8(3-4):279--292, 1992.

\bibitem{lin1992self}
Long-Ji Lin.
\newblock Self-improving reactive agents based on reinforcement learning,
  planning and teaching.
\newblock {\em Machine learning}, 8(3-4):293--321, 1992.

\bibitem{mnih2013playing}
Volodymyr Mnih, Koray Kavukcuoglu, David Silver, Alex Graves, Ioannis
  Antonoglou, Daan Wierstra, and Martin Riedmiller.
\newblock Playing atari with deep reinforcement learning.
\newblock {\em arXiv preprint arXiv:1312.5602}, 2013.

\bibitem{sallans2004reinforcement}
Brian Sallans and Geoffrey~E Hinton.
\newblock Reinforcement learning with factored states and actions.
\newblock {\em The Journal of Machine Learning Research}, 5:1063--1088, 2004.

\bibitem{sutton1998reinforcement}
Richard~S Sutton and Andrew~G Barto.
\newblock {\em Reinforcement learning: An introduction}.
\newblock MIT press, 1998.

\end{thebibliography}

\end{document}